%% file: main.tex
\documentclass[10pt]{article} %

\newif\ifarxiv
\arxivtrue

\usepackage[preprint]{rlj}

\usepackage{amssymb}            %
\usepackage{mathtools}          %
\usepackage{mathrsfs}           %
\usepackage{graphicx}           %
\usepackage{subcaption}         %
\usepackage[space]{grffile}     %
\usepackage{url}                %
\usepackage{lipsum}             %

\usepackage{amsmath}
\usepackage{enumitem}
\usepackage{amssymb}
\usepackage{graphicx}
\usepackage{bm}
\usepackage{theorem}
\usepackage{sidecap}
\usepackage{mathtools}

\usepackage[title]{appendix}
\usepackage{comment}
\usepackage{amsfonts}
\usepackage{dsfont}
\usepackage{hyperref}
\usepackage{lipsum}
\usepackage{dsfont,subcaption,float}
\usepackage{algorithm}
\usepackage{algorithmic}
\usepackage{multirow}

\usepackage{thm-restate}
\usepackage[size=small,labelfont=bf]{caption}
\usepackage{harpoon}%
\usepackage[normalem]{ulem}

\usepackage{colortbl} 
\usepackage{pgf} %
\usepackage{booktabs} %
\definecolor{lightblue}{HTML}{6699FF}

\usepackage[labelfont=bf, font=normalsize]{caption}
\newcommand{\algname}{\texttt{DRL$^2$}}
\newcommand{\eename}{\texttt{E2E}}

\newcommand{\gradientcella}[1]{%
  \pgfmathsetmacro{\value}{100 * (#1 - (-0.35) / (0.45 - (-0.35)))}
  \edef\colorvalue{\noexpand\cellcolor{lightblue!\value}}%
  \colorvalue #1%
}

\newcommand{\gradientcellb}[1]{%
  \pgfmathsetmacro{\value}{100 * (#1 - (-0.45) / (0.1 - (-0.45)))}
  \edef\colorvalue{\noexpand\cellcolor{lightblue!\value}}%
  \colorvalue #1%
}

\newcommand{\gradientcellc}[1]{%
  \pgfmathsetmacro{\value}{100 * (#1 - (-0.33) / (0.5 - (-0.33)))}
  \edef\colorvalue{\noexpand\cellcolor{lightblue!\value}}%
  \colorvalue #1%
}

\newcommand{\gradientcelld}[1]{%
  \pgfmathsetmacro{\value}{100 * (#1 - (-0.44) / (0.37 - (-0.44)))}
  \edef\colorvalue{\noexpand\cellcolor{lightblue!\value}}%
  \colorvalue #1%
}

\input{macros}

\title{An Empirical Study on the Power of Future Prediction in Partially Observable Environments}

\setrunningtitle{An Empirical Study on the Power of Future Prediction in Partially Observable Environments}

\author{Jeongyeol Kwon\textsuperscript{2,$\star$}, Liu Yang\textsuperscript{1,$\star$}, Robert Nowak\textsuperscript{2}, Josiah Hanna\textsuperscript{1}}

\emails{\{jeongyeol.kwon, liu.yang, rdnowak\}@wisc.edu, \ \{jphanna\}@cs.wisc.edu}

\affiliations{
$^{1}$\textbf{Department of Computer Science, University of Wisconsin-Madison, USA}\\
$^{2}$\textbf{Wisconsin Institute for Discovery, University of Wisconsin-Madison, USA} \\
\par %
$^\star$ equal contribution
}

\begin{document}

\maketitle  %

\begin{abstract}
Learning good representations of historical contexts is one of the core challenges of reinforcement learning (RL) in partially observable environments. 
While self-predictive auxiliary tasks have been shown to improve performance in fully observed settings, their role in partial observability remains underexplored. In this empirical study, we examine the effectiveness of self-predictive representation learning via future prediction--i.e., predicting next-step observations--as an auxiliary task for learning history representations, especially in environments with long-term dependencies.
We test the hypothesis that future prediction alone can produce representations that enable strong RL performance. To evaluate this, we introduce \algname, an approach that explicitly decouples representation learning from reinforcement learning, and compare this approach to end-to-end training across multiple benchmarks requiring long-term memory. Our findings provide evidence that this hypothesis holds across different network architectures, reinforcing the idea that future prediction performance serves as a reliable indicator of representation quality and contributes to improved RL performance.
\end{abstract}

\section{Introduction}
\input{Introduction}

\section{Preliminaries}\label{sec:prelim}

\input{Preliminaries}

\section{Decoupled Representation Learning and Reinforcement Learning}\label{sec:method}
\input{Methods}

\section{Experiments}\label{sec:exp}
\input{Experiments}

\section{Conclusion}
We study the empirical benefits of representation learning through future observation prediction for RL in partially observed environments. We demonstrate the high correlation between prediction performance and RL performance in several memory-intensive tasks. Furthermore, we consistently observe that auxiliary losses, often alone, can lead to good representations that significantly boost the efficiency of reinforcement learning. This highlights the need for continuing research on designing powerful auxiliary tasks to enhance reinforcement learning in partially observed environments.

There are several directions that may improve the base \algname~framework. First, our experiments employ only the minimalist implementation to test in all benchmarks; it would be interesting future work to investigate strategic exploration approaches in more complex environments with sparse rewards~\citep{saade2023unlocking, qiu2022contrastive, guo2022byol}. Second, the PSR learning framework assumes a set of core tests; 
while in most of our experiments, a one-step random action sufficed as the core test,
it would be an interesting future direction to automate the design of core tests in more challenging environments \citep{lambrechts2023informed, veeriah2019discovery, he2022reinforcement, jiang2021prioritized, lee2020predictive}. Finally, beyond RNNs and transformers, there is ongoing research that develops next-generation sequence models ({\it e.g.,} S4, S5) \citep{gu2021efficiently, smith2022simplified}; it would be worth investigating their power in more challenging environments where the prediction tasks themselves can be significantly more challenging.

\bibliography{main}
\bibliographystyle{rlj}

\newpage
\ifarxiv
\else
\beginSupplementaryMaterials
\fi

\input{Appendix}

\end{document}

%% file: macros.tex
\theoremstyle{plain}

\newcommand{\mA}{\mathcal{A}}	%
\newcommand{\mO}{\mathcal{O}}	%
\newcommand{\mT}{\mathcal{T}}	%
\newcommand{\mD}{\mathcal{D}}	%
\newcommand{\PP}{\mathds{P}}    %

\newcommand{\Exs}{\mathbb{E}}

\newcommand{\tssum}{\textstyle \sum}

\allowdisplaybreaks
\newif\ifarxiv

\newif\ifdraft
\drafttrue  %

%% file: Introduction.tex
Extensive research has explored reinforcement learning (RL) in fully observable environments, which are represented as Markov Decision Processes (MDPs). However, many real-world problems are partially observable, and there is a recent surge of interest in addressing more complex tasks where the environment state is only partially observable due to various factors such as missing information or noisy measurements. In such situations, the agent must consider the entire history to make optimal decisions~\citep{smallwood1973optimal}. %

To succeed in partially observable environments, one natural approach is to learn a \emph{good representation of history}, effectively transforming the problem into an MDP.
Expanding on this approach, recent works leverage sequence models, such as recurrent neural networks (RNNs)~\citep{hochreiter1997long} or transformers~\citep{vaswani2017attention}, to learn history representations. These models are then integrated with RL techniques designed for fully observable MDPs, such as policy gradient methods~\citep{schulman2017proximal} and temporal-difference learning~\citep{sutton2018reinforcement}, and trained end-to-end~\citep{wierstra2007solving, hausknecht2015deep, meng2021memory}.
However, when optimal decisions depend on long-term histories, the end-to-end approaches often struggle to learn good representations of long histories, possibly because policy gradients alone do not provide a sufficiently strong learning signals~\citep{morad2022popgym}. 

As an alternative to learning representations solely through RL signals, several auxiliary tasks have been proposed. One promising approach is to employ {self-predictive representation learning}~\citep{ni2024bridging}, which learns the representation of histories through (self-)supervised learning to \emph{predict future events} in observation or latent space. The appeal of future prediction as an auxiliary task stems not only from empirical evidence~\citep{oh2015action, oh2017value, jaderberg2016reinforcement, oord2018representation, hefny2018recurrent, igl2018deep, gregor2019shaping, gelada2019deepmdp, anand2019unsupervised, guo2020bootstrap, yu2020meta, schwarzer2020data, castro2021mico, khetarpal2024unifying, stooke2021decoupling}, but also from the theory of predictive state representation~\citep{littman2002predictive, singh2004predictive, boots2011closing, hefny2015supervised, uehara2022provably, liu2023optimistic}. 

Despite the extensive literature on future prediction representation learning methods, their effectiveness remains uncertain: whether practical advancements come from genuine methodological innovations or merely from improved implementation engineering, as concluded in \cite{ni2022recurrent}. Previous works often focus on benchmarks like noisy versions of Atari~\citep{bellemare2013arcade} and MuJoCo~\citep{todorov2012mujoco}, where tasks can be solved using short-term memory alone. Moreover, existing methods frequently fail to outperform well-tuned end-to-end training~\citep{ni2022recurrent, grigsby2023amago} on benchmarks that require long-term memory~\citep{morad2022popgym}.

To this end, this paper empirically investigate the following question: 
\vspace{-0.5em}
\begin{center}
    \emph{Can future observation prediction tasks produce learned representations that support and improve RL in partially observable environments?} 
\end{center}

To investigate this question, we fully \underline{\texttt{D}}ecouple the history \underline{\texttt{R}}epresentation \underline{\texttt{L}}earning from \underline{\texttt{R}}einforcement \underline{\texttt{L}}earning, introducing our approach, \algname. Specifically, \algname~learns history representations through a decoupled future observation prediction task. We then compare \algname~to end-to-end training (\eename), where the history representation and policy network are jointly learned through the RL signal.
We summarize our key findings below:
\begin{itemize}
    \item In pure-memorization tasks, future prediction directly leads to good representations for reinforcement learning. (Section \ref{subsec:long_term_memory}). 
    \item In tasks requiring long-term reasoning, future prediction significantly improves the performance of reinforcement learning (Section \ref{subsec:complex_reasoning}).
    \item In tasks with sparse rewards, future prediction yields faster and stable convergences of reinforcement learning (Section \ref{subsec:temp_cred_assign}).
    \item In tasks where short-term memory suffices, future prediction may slightly underperform compared to end-to-end training in terms of sample complexity (Appendix \ref{subsec:shot_term_memory}).
\end{itemize}
Importantly, our analysis is not tied to a particular architecture: the principles we present hold true for a range of RNNs and transformers. 
Furthermore, our experiments strongly advocate decoupling the representation learning phase from reinforcement learning for both sample and computational efficiency when solving POMDPs with long historical dependencies. %

We perform our empirical analysis on three challenging partially observable benchmarks: (1) POPGym \citep{morad2022popgym}, (2) mujoco simulators with partial masking \citep{todorov2012mujoco}, and (3) credit assignment benchmarks \citep{ni2022recurrent}. 
Our goal is to highlight the strong correlation between future prediction performance and RL performance, measured by cumulative returns, in memory-intensive POMDP benchmarks. This provides new insights into the role of self-predictive representation learning in reinforcement learning. We believe this understanding could pave the way for more generalized and scalable RL methods in partially observable environments.

The remainder of the paper is organized as follows. In Section~\ref{sec:prelim}, we formalize the problem in the framework of predictive state representation. Subsequently, we introduce our decoupled representation learning method (\algname) in Section~\ref{sec:method}. The experimental results are presented in Section~\ref{sec:exp}. %

\section{Related Work}
\label{sec:related_work}
\input{RelatedWork}

%% file: RelatedWork.tex
RL in partially observed environments is an active area of research that also arises in many subtopics \citep{duan2016rl, zintgraf2019varibad, kirk2021survey, kwon2021rl}. Due to the vast and growing literature, we only review subareas of RL that are most relevant to us.

\paragraph{End-to-End Training vs Auxiliary Losses.} End-to-end training has proven to be a compelling choice, with recent work showcasing its potential when combined with carefully engineered architectures and hyperparameters \citep{ni2022recurrent, ni2023transformers, grigsby2023amago}. The philosophy of end-to-end training is to resolve two major issues simultaneously within a single optimization pass: (a) learning good history representations and (b) learning a good policy with the learned representation. However, signals from a single optimization pass ({\it e.g.,} policy gradients or temporal difference updates) may not always lead to representations that accurately reflect the true state of the environment, especially in the early phase of the training. We investigate this phenomenon in greater details in Section \ref{subsec:long_term_memory} and \ref{subsec:complex_reasoning}. %

\paragraph{Representation Learning in RL.} Even in fully observable domains, the significance of representation learning cannot be overstated, as it plays a crucial role in compressing high-dimensional observations to enhance overall RL performance. Recent research has introduced a variety of auxiliary losses, some of which are driven by predicting future observations \citep{khetarpal2024unifying}, rewards \citep{oh2017value, liu2019sequence}, and other unsupervised signals \citep{de2018integrating, oord2018representation, schwarzer2020data, guo2020bootstrap, stooke2021decoupling, qiu2022contrastive, konan2023contrastive}. There is a growing body of research \citep{voelcker2024does,ni2024bridging} investigating the benefits of auxiliary prediction tasks more systematically. While their conclusions also support the power of representation learning in reinforcement learning, their demonstration is mostly focused on MDPs with distracting observations, and less on memory-intensive benchmarks, with a stronger advocation toward the {\it latent state} prediction. We focus on observation prediction as an auxiliary task since the performance is always verifiable from data without ambiguity, whereas the latent state prediction loss may not tell much by itself unless the learned latent state is guaranteed to be meaningful.

\paragraph{Decoupled Representation Learning from Reinforcement Learning.} Most of these studies have intertwined the RL losses with auxiliary losses by incorporating them in parallel and training the entire system through combining both losses, limiting our understanding of the true impact of prediction tasks. One exception is the work of \citet{stooke2021decoupling}, where the authors successfully decoupled representation learning from reinforcement learning, achieving results at least on par with the best end-to-end training outcomes. Our work follows a similar spirit, demonstrating that effective representation learning of long histories can be achieved solely through representation learning. As a result, our overall RL performance can match and often outperform that of well-tuned end-to-end RL while providing deeper insights into the underlying training dynamics.

\paragraph{Architectures for POMDPs.} While RNN or Long Short-Term Memory (LSTM) \citep{hochreiter1997long} have been dominant in the past \citep{hausknecht2015deep, ni2022recurrent}, Transformer architectures have gained significant attention for their remarkable performance in processing sequential data \citep{radford2019language}. There has been growing excitement within the RL community surrounding the potential of transformers as a universal solution for sequential decision-making \citep{chen2021decision, zheng2022online, konan2023contrastive, ni2023transformers}. More recently, architectures inspired by state-space based modeling are also gaining more interests \citep{gu2021efficiently, smith2022simplified, luis2024uncertainty}. Our architecture choices are solely based on the capacity of each architecture to learn the prediction task, as we demonstrate the high correlation between representation learning and reinforcement learning performance empirically. %

\paragraph{Other Task Formulations in POMDPs.} One line of work considers model-based approaches within the framework of partially observable MDPs (POMDPs) \citep{igl2018deep, azar2019world, zintgraf2019varibad, gregor2019shaping, buesing2018learning, chen2022flow, avalos2023wasserstein, wang2023learning}. Although their problem settings can be rephrased as learning latent representations of histories (known as belief states \citep{pineau2006anytime}), learning a POMDP model necessitates the inference of unobservable hidden states. This makes the training objective much more difficult to optimize unless aided by access to internal states during training.

%% file: Preliminaries.tex
We formalize the problem in the framework of predictive state representation (PSR) \citep{littman2002predictive, singh2004predictive}, a versatile framework that bypasses the strict modeling requirement of hidden states in POMDPs \citep{smallwood1973optimal}. PSR represents the underlying state as the probability of future events, without making any assumptions about hidden dynamics. This makes the concept of PSR particularly well suited for model-free approaches.

\paragraph{Predictive State Representation.} A PSR consists of the tuple $(\mA, \mO, \PP, r)$ action space $\mathcal{A}$; observation space $\mathcal{O}$; a probability distribution over all possible histories $\PP: (\mathcal{O}\times \mathcal{A})^*  \rightarrow [0,1]$ %
where $\PP(\cdot | o_1, a_1, ..., a_{t-1})$ is the conditional probability of $o_t$ given a sequence of arbitrary length $t$ of observations $o_{1:t} := (o_1, ..., o_t)$ and a sequence of actions $a_{1:t-1} := (a_1, ..., a_{t-1})$ is executed; and $r_t$ is an instantaneous observable reward. %
We consider episodic environments with finite time-horizon $H > 0$.

The goal of RL in a PSR is to learn a history-dependent policy $\pi: (\mO \times \mA)^* \times \mO \rightarrow \Delta(\mA)$ that maximizes the expected cumulative reward in an episode:
\begin{align}
    J(\pi) = \Exs \left[\tssum_{t=1}^H r_t \ \big| a_t \sim \pi(\cdot | h_t) \right], \label{eq:reward_function}
\end{align}
where $h_t = (o_1, a_1, ..., a_{t-1}, o_t)$ is a given history at the $t^{\texttt{th}}$ time step, and $\Exs$ is an expectation over histories generated when taking actions according to policy $\pi$.

\paragraph{Future Prediction and Core Tests.} In a PSR, we represent a history through prediction on a finite set of {\it core tests} \citep{littman2002predictive}. Specifically, let
$\mT := \{ (a_{1:k}^q, o_{1:k}^q)\}_{q}$ where $a_{1:k}^q$ is a sequence of test actions and $o_{1:k}^q$ is a sequence of future observations. Then, the predictive state of a history $h_t$ is given by
\begin{align*}
    \PP (o_{1:k}^q \| \textbf{ do } a_{1:k}^q | h_t) = \psi(o_{1:k}^q | \phi(h_t), a_{1:k}^q), 
\end{align*}
where $\PP (o_{1:k}^q \| \textbf{ do } a_{1:k}^q | h_t)$ is the probability of observing $o_{1:k}^q$ after executing $a_{1:k}^q$ following observing history $h_t$, and $\phi, \psi$ are history summarizing and future prediction models, respectively. 
We mention that instead of a sequence of raw future observations, one may also learn the encoding of the observation sequence \citep{guo2020bootstrap, guo2022byol}; we do not pursue that direction in this work.

As we describe in the next section, representation learning via future prediction can be cast as self-supervised learning \citep{hefny2015supervised}, without the need to learn dynamics in latent states. Our work aims to demonstrate that with sufficiently accurate future prediction on core tests, we can view $\phi(h)$ as a good representation of $h$.

%% file: Methods.tex
\begin{SCfigure}[50]
    \centering
    \includegraphics[width=0.49\textwidth]{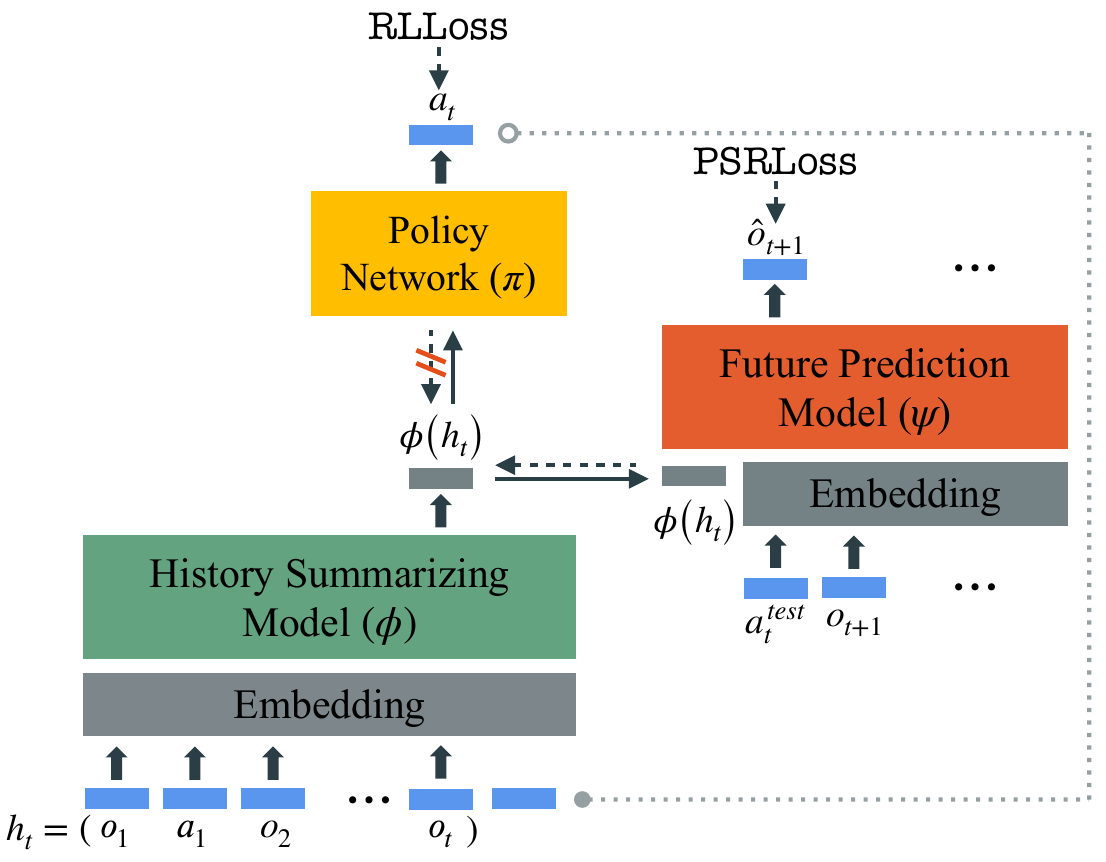}
    \caption{\small \emph{Decoupled representation learning and reinforcement learning}. RL in partially observed environments can be decomposed into: (a) learning a summarized state representation using the history summarizing model ($\phi$), and (b) optimizing the policy network ($\pi$) based on this representation.
    Instead of training these components end-to-end, we decouple them by supervising representation learning through a co-trained future prediction model ($\psi$), which guides the learning of $\phi$ via future representation signals (\texttt{PSRLoss}). The policy network is updated separately using the reinforcement learning signal (\texttt{RLLoss}) while being conditioned on the decoupled representation from $\phi$.
    }
    \label{fig:architecture}
\end{SCfigure}

Our goal is to evaluate how effectively future prediction can facilitate learning representations of long histories. To do so, we completely decouple representation learning from reinforcement learning, and learn the representation purely based on the future prediction task. As a result, our training pipeline alternates between the two following tasks: %
\begin{enumerate}
    \item \emph{(Predictive State Representation Learning)} The history sequence $h_t = (o_1, a_1, ..., o_t)$ is input into the history summarizing model $\phi(\cdot)$, yielding summary statistics $\phi(h_t)$. Subsequently, $a_t$ is selected from $\pi^{\texttt{test}}$, then input into the future prediction model, along with $\phi(h_t)$, to estimate future observation probabilities.
    The optimization objective is to minimize the \texttt{PSRLoss}:
    \begin{align}
        &\texttt{PSRLoss}(\psi, \phi; \mD)  := \underset{(h,a_{1:k},o_{1:k})}{\Exs_{\mD}}  \left[\ell (o_{1:k}; \psi(o_{1:k} \textbf{ do } a_{1:k} | \phi(h))) \right], 
    \end{align}
    where $l(\cdot; \cdot)$ is a loss function -- we use cross-entropy (CE) for discretized observations, and mean-squared error (MSE) for continuous observations. $\mD$ is a trajectory dataset.
    Both sequence models $(\phi, \psi)$ are updated in the representation learning phase.     
    \item \emph{(Reinforcement Learning)} In this phase, the history summarizing sequence model is frozen, and trajectories are generated by leveraging the learned representation model $\phi$ alongside the policy network $\pi: (\mathcal{O}\times\mathcal{A})^* \rightarrow \Delta(\mA)$. Then the policy network $\pi$ is updated by optimizing the $\texttt{RLLoss}(\pi; \phi, \mD)$, which maximizes the cumulative returns with a history-to-representation mapping $\phi$: 
    \begin{align}
        \Exs \left[\tssum_{t=1}^H r_t \ \big| a_t \sim \pi(\cdot | \phi(h_t)) \right]. \label{eq:rl_loss}  
    \end{align}
    In our implementation, we use the soft actor-critic formulation \citep{haarnoja2018soft, christodoulou2019soft} using the same trajectory dataset $\mD$, since off-policy methods tend to be more sample efficient and stable.  
\end{enumerate}

\paragraph{Training Algorithms.} We summarize the training procedure in Algorithm \ref{algo:interleaving_PSR}. Training alternates between two phases: data generation and model training, each running for $T_{gen}$ and $T_{tr}$ iterations. 
\begin{enumerate}
    \item \emph{Data Generation Phase}: The agent collects data using the current policy network $\pi$, except for one randomly chosen timestep where it takes a test action instead (see Appendix~\ref{appendix:env} for details).
    \item \emph{Model Training Phase}: Representation learning and reinforcement learning updates are interleaved with step sizes $\alpha$ and $\beta$. In \algname, we block gradients from $\pi$ to $\phi$ to ensure decoupled representation learning.
\end{enumerate}
For a fair comparison, we implement the end-to-end (\eename) training follows the same training procedure as Algorithm~\ref{algo:interleaving_PSR}, except that during the model training phase, we set $\alpha=0$ and update $(\phi, \psi)$ solely through \texttt{RLLoss}.

\begin{algorithm}[h!]
\resizebox{0.9\textwidth}{!}{
\begin{minipage}{\textwidth}
    \caption{Decoupled Representation Learning and Reinforcement Learning (\algname)}
    \label{algo:interleaving_PSR}
    \begin{algorithmic}[1]
        \STATE{\textbf{Inputs:} step sizes $\alpha, \beta$, tunable inner steps $T_{tr}, T_{gen}$}
        \STATE{(Optional) Burn-In Phase: Generate an initial dataset $\mD$ with random policy and pre-train $(\psi, \phi)$}
        \FOR{$i \ge 1$} 
            \STATE{\color{blue}{\# \emph{Data Generation Phase}: Collecting trajectories}}
            \FOR{$j \in [T_{gen}]$}
                \STATE{$t \sim \text{Unif}([H])$, add a new trajectory sample to $\mD$ with $\pi \circ_t a^{\texttt{test}}_t \circ_{t+1} \pi$}
            \ENDFOR
            \STATE{\color{blue}{\# \emph{Model Training Phase}: Interleaved Representation Learning and Reinforcement Learning}}
            \FOR{$j\in [T_{tr}]$}
                \STATE{$(\psi, \phi) \leftarrow (\psi, \phi) - \alpha \hat{\nabla}_{(\psi, \phi)} \texttt{PSRLoss} (\psi, \phi; \mD)$}
                \STATE{$\pi \leftarrow \pi - \beta \hat{\nabla}_{\pi}  \texttt{RLLoss} (\pi; \phi, \mathcal{D})$}
            \ENDFOR
        \ENDFOR
    \end{algorithmic}
\end{minipage}
}
\end{algorithm}

\paragraph{Architecture.} 
Unless stated otherwise, we use a transformer for the history summarizing model $\phi$ and a GRU for the future prediction model $\psi$. The transformer follows the NanoGPT implementation\footnote{\url{https://github.com/karpathy/nanoGPT}}.
For the policy network, we employ the soft actor-critic (SAC)~\citep{haarnoja2018soft} for continuous actions, and SACD for discrete actions~\citep{christodoulou2019soft}. 
The projection from input states and actions to the embedding is shared by the history summarizing model and the future prediction model. The remaining details can be found in Appendix~\ref{appendix:model}.

The overall model and training pipeline are depicted in Figure \ref{fig:architecture}. It is worth noting that we adhere to the {\it minimalist approach} to architecture design and base algorithms.  This is primarily for focusing on the true impact of future prediction; yet, this simple implementation can be sufficient in many challenging benchmarks that require the ability to decide the optimal actions from long histories.

%% file: Experiments.tex
We evaluate our framework, \algname, on selected partially observed environments, including (1) the GridWorld environment, (2) the memory-intensive tasks and complex reasoning tasks in POPGym benchmark~\citep{morad2022popgym}, (3) temporal credit assignment tasks such as Delayed Catch~\citep{Raposo2021SyntheticRF} and Dark Key-to-Door~\citep{Laskin2022IncontextRL}, and (4) MuJoCo~\citep{todorov2012mujoco}. 
In reporting the returns, we compute the average cumulative rewards over the most recent 5,000 episodes. A summary of all the environments we have investigated is presented in Appendix~\ref{appendix:exp_results}. 
Throughout the experiments, we ran 3 different trials and plotted the average with a shaded area indicating the mean-std error.

It is important to note that our goal is not to compete for state-of-the-art benchmark performance. Instead, we study
\emph{whether future prediction accuracy have a meaningful connection to reinforcement learning performance in partially observed environments?} 

\subsection{Preliminary Study on State-Based vs. Stateless Agents}\label{subsec:gridworld}
We start with a preliminary experiment in a noisy observation environment--GridWorld.
This preliminary setup enables us to compare (1) a memoryless agent, (2) the end-to-end (\eename) training, and (3) our method \algname. 
The stateless agent is trained using a minimal soft actor-critic (SAC) implementation. 
For both \algname\ and \eename, during the burn-in phase, we pre-fill the buffer with 5k trajectories generated by the untrained, randomly initialized model.

\begin{minipage}{0.4\textwidth}
    \centering
    \includegraphics[width=0.75\textwidth]{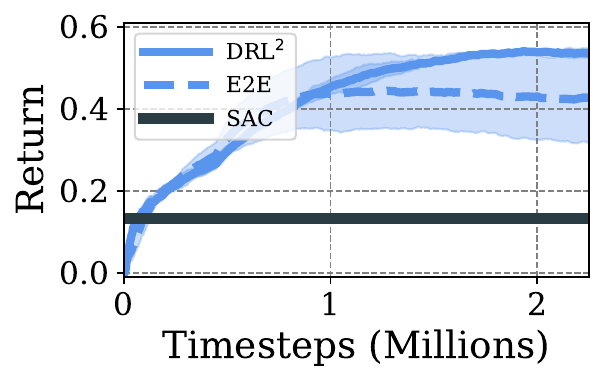} 
    \vspace{-4mm}
    \captionof{figure}{\small \emph{GridWorld environment with noise in observations.} 
    }
    \label{fig:grid_world}
\end{minipage}
\hfill
\begin{minipage}{0.59\textwidth}
    As shown in Fig.~\ref{fig:grid_world}, the stateless agent achieves an average return of around 0.12, while the \algname~framework outperforms this baseline and achieves performance comparable to the \eename~framework. These findings suggest that in partially observed settings, the \algname\ and \eename\ methods are more effective than the stateless agent. In subsequent environments, our discussion and analyses will focus exclusively on the \algname\ and \eename\ methods.
\end{minipage}

\subsection{Sequential Model Choices on Memorization Benchmarks} \label{subsec:long_term_memory}

{Encoding long-term memory in the representation} is one of the challenges in the partially observed environments.   
The capacity for long-term recall is contingent on both the learned representation and the memory capacity of the sequential models.
In this section, we examine how the choice of \emph{history summarizing model} affects \eename~and \algname~in pure long-term recall tasks, using RepeatPrevious from POPGym~\citep{morad2022popgym}. As mentioned before, the future prediction model is a GRU.
We compare transformers (TF)~\citep{ni2023transformers}, using the NanoGPT implementation, with Gated Recurrent Units (GRUs)~\citep{Cho2014LearningPR}, a widely used and strong baseline~\citep{grigsby2023amago, morad2022popgym}. Additionally, Amago~\citep{grigsby2023amago} is a state-of-the-art \eename~benchmark that introduces modifications to transformers for more effective end-to-end training. We also adopt this modified transformer implementation\footnote{\url{https://github.com/UT-Austin-RPL/amago/blob/main/amago/nets/transformer.py}} as the history summarizing model, referring to it as Amago-GPT.

\begin{figure}[h!]
    \centering
    \ifarxiv
    \includegraphics[width=0.5\textwidth]{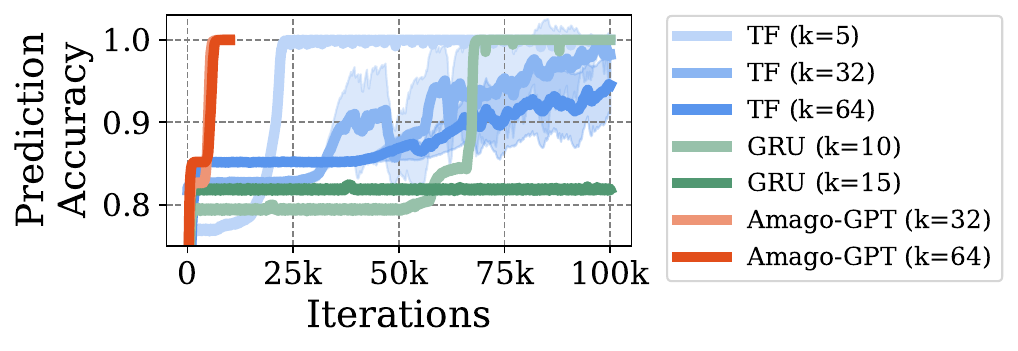}\else\includegraphics[width=0.44\textwidth]{Figures/Repeat_Previous_Burn_in.pdf}
    \fi
    \includegraphics[width=0.55\textwidth]{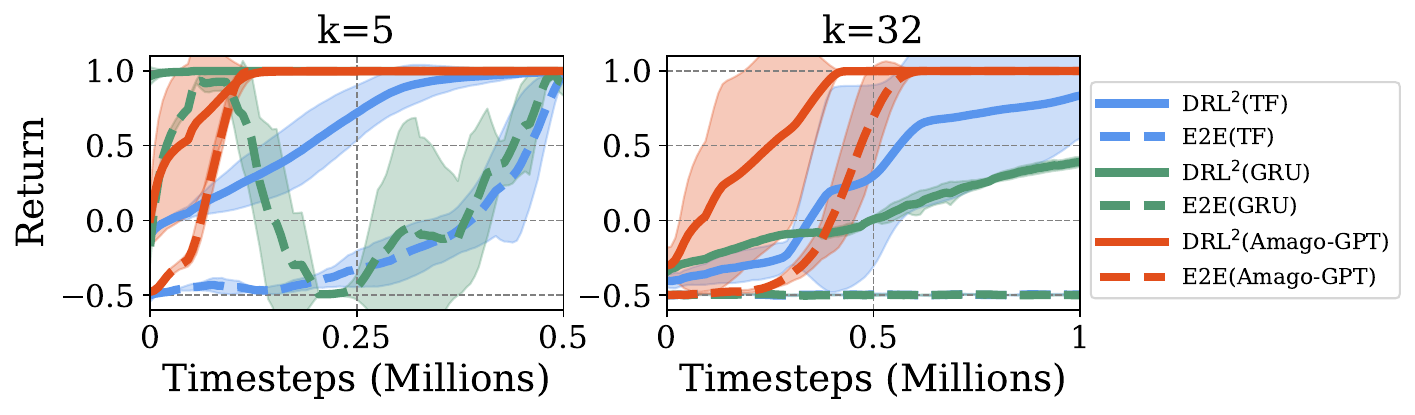}
    \vspace{-2em}
    \caption{\small \emph{Future prediction performance during the burn-in phase (left) and RL performance (right) across different sequential models.} In the RepeatPrevious experiment, we compare prediction accuracy using transformers (TF), GRUs, and Amago-GPT as the history summarizing model across varying difficulty levels defined by k. Amago-GPT consistently achieves fast and stable convergence across different k, while TF and GRUs struggle at higher k values. In these cases, \eename~fails to learn recall, whereas \algname~successfully does.
    }
    \vspace{-1.4em}
    \label{fig:repeat_previous}
\end{figure}

\paragraph{Prediction Performance During Burn-in and Its Relationship to RL Performance.} 
In RepeatPrevious, the agent must remember the suit dealt k steps ago, and each observation includes a reward indicator showing whether the recalled suit was correct.
In \algname, predicting this indicator can be trained and evaluated during the burn-in phase using trajectories from a random policy. This allows us to isolate the influence of trajectory quality from the learned policy and focus solely on the architecture’s role in future prediction performance. Fig.~\ref{fig:repeat_previous} (\emph{left}) shows prediction accuracy across different sequential models in the history summarizing model. Transformers achieve near-perfect accuracy even at k=64, while GRUs struggle beyond k>10. Both sequential models require more training as k increases, highlighting the challenge of learning long-term dependencies. Notably, Amago-GPT converges quickly and stably even at k=64, suggesting its advanced performance may come from the enhanced memory capacity due to architectural modifications.
Burn-in phase performance strongly correlates with final RL performance. As shown in Fig.~\ref{fig:repeat_previous} (\emph{right}), while all sequential models reach optimal returns when $k=5$, only Amago-GPT successfully recalls the correct card at $k=32$ under \eename. In contrast, \algname~consistently achieves non-random performance across all models, suggesting that it enhances weaker sequential architectures, improving their ability to handle memory-intensive tasks.

\begin{SCfigure}[50][h]
    \centering
    \ifarxiv\includegraphics[width=0.6\textwidth]    {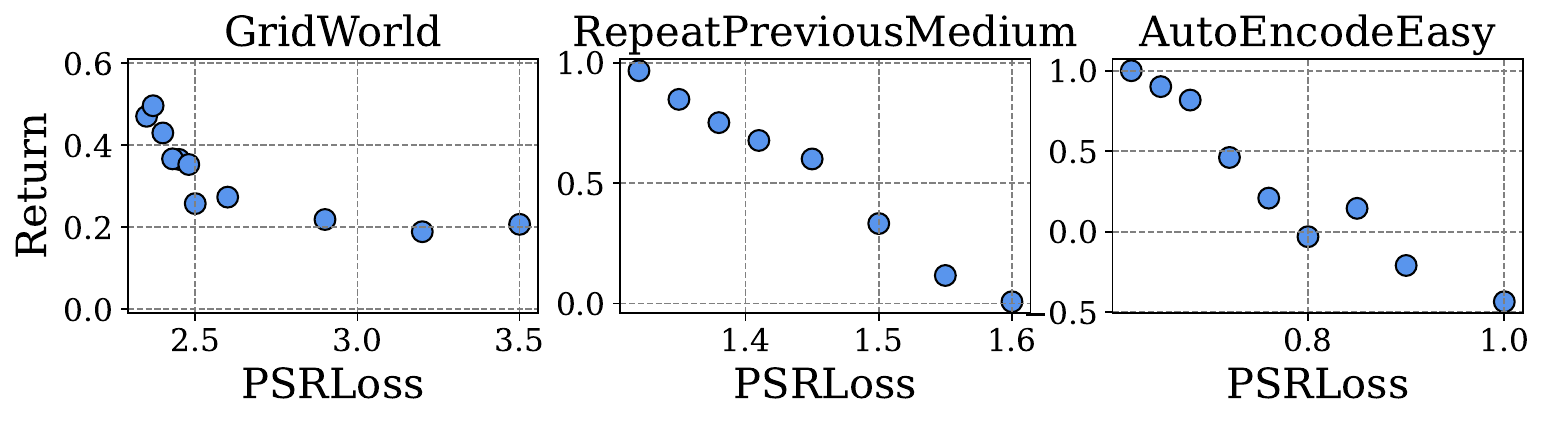}
    \else\includegraphics[width=0.6\textwidth]{Figures/psr_vs_return_gridworld_repeat_and_autoencode.pdf}
    \fi
    \vspace{-0.5em}
    \caption{\small \emph{Correlation between the \texttt{PSRLoss} and the average return of the model.} A consistent correlation is observed between future prediction loss (\texttt{PSRLoss}) and returns.}
    \label{fig:psr_vs_return_repeat_autoencode}
\end{SCfigure}

\paragraph{Effects of Prediction Performance on RL Performance.}
Decoupling representation learning in \algname\ is effective, but how well does the PSR loss reflect representation quality? To investigate, we first train the history summarizing model to a set PSR loss, then freeze it and train the policy network separately on top of the representations produced by the frozen history summarizing model. We evaluate this in GridWorld (Section~\ref{subsec:gridworld}) and two memory-intensive environments, RepeatPrevious and AutoEncode. 
Since the history summarizing model is frozen during policy training, any observed improvement in RL performance stems solely from the quality of representations learned through future prediction, rather than adaptation during RL training. As shown in Fig.~\ref{fig:psr_vs_return_repeat_autoencode}, agents with more precise future predictions—reflected in lower PSR loss—tend to achieve higher average returns, confirming that PSR loss is a reliable indicator of representation quality.

\vspace{-0.8em}
\subsection{Long Sequential Games with Prediction}
\label{subsec:complex_reasoning}

We further investigate more challenging long sequential game environments from PopGym \citep{morad2022popgym}: Battleship and Minesweeper (details on environment descriptions are provided in Appendix~\ref{appendix:env}). In these environments, an agent must infer long-term correlations between historical contexts and future observations rather than simply memorize and recall certain events.

\begin{table}[h]
\centering
\resizebox{0.8\textwidth}{!}{
\begin{tabular}{l|ccc|ll}
    \toprule
    & \eename@15M & \algname@15M & Best Baselines & \algname~Converged & Amago Reproduced\\
    \midrule
    Minesweeper (Medium) & \gradientcella{-0.35} & \gradientcella{ 0.20} & \gradientcella{ 0.33}\textsuperscript{\emph{ Amago}} & \gradientcella{0.45} @ 45M & \gradientcella{ 0.45} @ 45M \\
    \hline
    Minesweeper (Hard) & \gradientcellb{-0.45} & \gradientcellb{-0.32} & \gradientcellb{-0.20}\textsuperscript{\emph{PopGym}} & \gradientcellb{0.10} @ 125M & \gradientcellb{ 0.03} @ 125M \\
    \hline
    Battleship (Medium) & \gradientcellc{-0.33} & \gradientcellc{-0.15} &  \gradientcellc{-0.21}\textsuperscript{\emph{ Amago}} & \gradientcellc{0.50} @ 75M & \gradientcellc{-0.01} @ 75M \\
    \hline
    Battleship (Hard) & \gradientcelld{-0.44} & \gradientcelld{-0.42} & \gradientcelld{-0.37}\textsuperscript{\emph{PopGym}} & \gradientcelld{0.37} @ 180M & \gradientcellb{-0.07} @ 180M \\
    \bottomrule
\end{tabular}
}
\vspace{-0.8em}
\caption{\emph{Improved performance via \algname~in long-sequence high-complexity games.} \algname~achieves notable performance improvements over established baselines in both medium and hard configurations of Minesweeper and Battleship.}
\label{table:long_sequence_games}
\vspace{-0.8em}
\end{table}

We compare \algname~against \eename~and the best reported baseline. To complement this comparison, we reproduce the performance of the strong baseline Amago~\citep{grigsby2023amago}, which has also been evaluated on these environments, and run it for either the same number of timesteps we report or until convergence. When reporting the reproduced Amago performance, we favor Amago by reporting its highest observed average return. Additionally, we note the difference in how trajectories are collected. Amago continuously collects trajectories, resetting the environment immediately upon termination. In contrast, we collect full episodes, and if an episode ends early, we pad the remaining steps with dummy tokens, which still count toward the total collected timesteps. As a result, our reported timesteps include some padding overhead.

The selected results, shown in Table~\ref{table:long_sequence_games}, demonstrate that \algname~is capable of achieving nearly-optimal returns in these environments, which has not been demonstrated in the prior works with on- and off-policy~\eename~methods~\citep{grigsby2023amago, morad2022popgym}. 
We note that our goal is not to claim that the strong baseline, Amago, fails at this task, as it may achieve better performance with further tuning. Instead, we aim to demonstrate that our \algname~method--learning representations from future prediction alone--can eventually conquer these tasks given a sufficient number of episodes.
More experimental results on these challenging long-sequential game environments can be found in Appendix~\ref{appendix:exp_results}.

\subsection{Temporal Credit Assignment}\label{subsec:temp_cred_assign}
Additionally, we study the sparse-reward temporal credit assignment tasks: Delayed Catch~\citep{Raposo2021SyntheticRF} and Dark Key-to-Door~\citep{Laskin2022IncontextRL}. In the delayed catch environment, the reward is not revealed until the end of multiple catches. In the dark key-to-door environment, the agent will only receive reward 1 if it finds an invisible key, then open an invisible door, otherwise the reward remains to be 0. 

\begin{figure}[h!]
    \centering
    \includegraphics[width=0.23\linewidth]{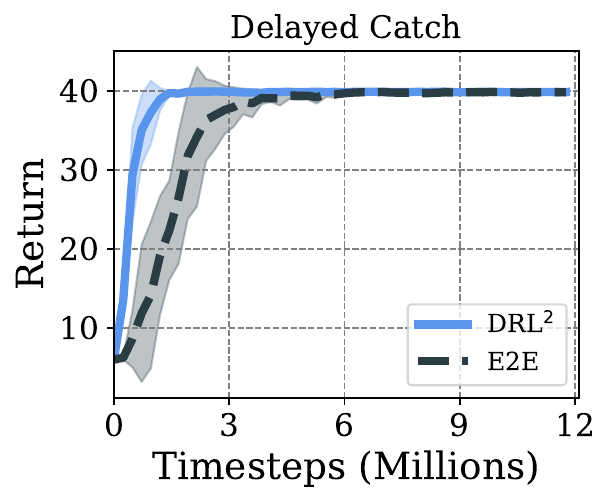}
    \includegraphics[width=0.23\linewidth]{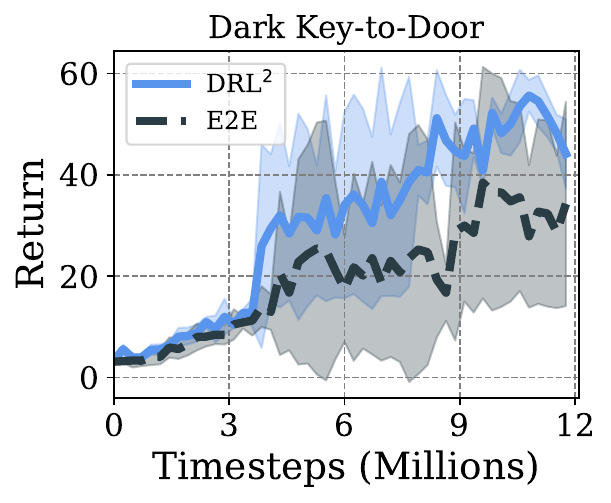}
    \includegraphics[width=0.29\linewidth]{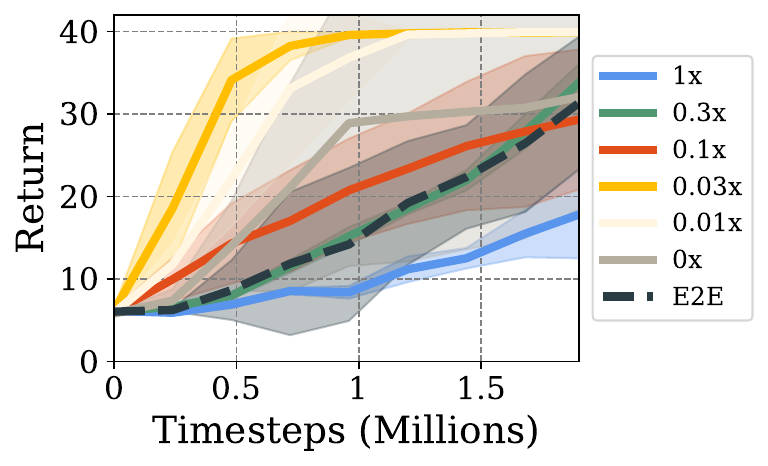}
    \vspace{-1em}
    \caption{\small \emph{Temporal credit assignment tasks performance (left), and effects of the update ratio between PSR and RL (right).} (Left) \algname~demonstrate better convergence compared to \eename. (Right) We evaluate the impact of varying update ratios between PSR and RL in the delayed catch environment, and compare its performance with the E2E method. Here, $0.1$x means that for every one update step of RL, PSR is updated only 0.1 times on average, i.e., once every 10 RL updates.}
    \label{fig:temp_cred_assign}
    \vspace{-1.5em}
\end{figure}

We use the implementation from AMAGO\footnote{\url{https://github.com/UT-Austin-RPL/amago}}\citep{grigsby2023amago} and compare the performance of \eename\ with \algname. The results are presented in Figure~\ref{fig:temp_cred_assign} (left). As shown in the figure, \algname\ consistently outperforms \eename, achieving the maximum reward more quickly.

Additionally, we analyze how the update ratio between RL and PSR affects learning in the Delayed Catch environment, with results shown in Figure~\ref{fig:temp_cred_assign} (right). 
As observed, the most effective update ratio is $0.03:1$ (PSR to RL), aligning with the bilevel optimization structure of PSR learning~\citep{kwon2024complexity}. Under this setting, \algname~achieves faster RL convergence compared to \eename, indicating that a slower PSR update provides a more stable training signal, benefiting overall RL performance. 
While this slower PSR update holds for most benchmarks we investigated, it may not always be optimal (Table~\ref{table:hyper_param_table} presents hyperparameter choices across different benchmarks). Interestingly, in this task, detaching and freezing representation learning alone can also induce performance comparable to \eename, but with high variance, suggesting that the task relies less on the learned representation.

\section{Discussion}

We have consistently observed that when the effective historical length is long, decoupling the representation learning from reinforcement learning shows more stable and faster convergence than the end-to-end training. We conjecture that loss signals from reinforcement learning objectives ({\it e.g.,} actor-critic losses) typically have much higher variances when historical contexts get longer, while representation (prediction) learning tasks are guided by supervised losses with lower variance. We leave it as future work to investigate more complex observations such as images or sentences in large-scale tasks. Below, we briefly discuss additional detailed points that can be useful in practice.

\paragraph{Parameter/Architecture Tuning.} It is worth noting that we adhere to the {\it minimalist approach} to architecture design and base algorithms, as long as architectures can conquer future prediction tasks conditioned on most historical context. On the algorithmic side, the only hyperparameters that we have tuned are the learning rates: $\alpha, \beta$ and the number of inner steps $T_{tr}, T_{gen}$ in Algorithm \ref{algo:interleaving_PSR}. We conclude that this minimalist implementation can be a strong baseline for many existing benchmarks with proper tuning of learning rates. In more complex environments, advanced architecture and auxiliary task designs may become crucial for learning good representations.

\paragraph{Recommendations for Practitioners.} To conclude our experiments, our recommendation of procedures to decide the implementation is the following: 
\begin{enumerate}
    \item \emph{Assess memory needs}. If short-term memory suffices, \eename~with simple RNNs should perform well. However, if longer-term memorization and reasoning are required, \eename~may struggle or require extensive hyperparameter tuning. In such cases, designing effective prediction losses as an auxiliary task can help.
	\item \emph{Key hyperparameters to tune}. In \algname, selecting the right ratio between $T_{tr}$, $T_{gen}$, and step sizes $\alpha$ and $\beta$ (as defined in Algorithm~\ref{algo:interleaving_PSR}) is important. We recommend starting with \eename’s settings and iteratively adjusting these parameters while comparing RL performance against \eename.
	\item \emph{Decouple training for failure diagnosis}. Decoupling representation learning from reinforcement learning can help diagnose challenges in each environment and improve sample efficiency. This includes verifying whether future prediction has converged and identifying cases where RL performance is stuck in a suboptimal state.
\end{enumerate}

%% file: Appendix.tex
\section{Details of Partially Observed Benchmarks}\label{appendix:env}
Below are the details of the partially observable environments we have explored in our benchmark.

\paragraph{Preliminary Environment: GridWorld.}
The GridWorld environment consists of a 7×7 discrete grid, where an agent must reach an unknown target location within a fixed time horizon of 9 steps. The target is randomly assigned at the start of each episode but remains hidden from the agent until it reaches the goal. The agent can move in four cardinal directions (up, down, left, right), but movements beyond grid boundaries result in no change in position. At each step, the agent receives three observations: (1) its current location (x, y), (2) a binary distance indicator that signals whether the last action reduced its distance to the target, which is correct with 90\% probability, and (3) a random noise observation uniformly sampled from $[0,1]$. The agent only obtains a reward if it successfully reaches the target within the time horizon.

\paragraph{Memory-Intense Environment.} 
\begin{enumerate}
    \item \emph{RepeatPrevious}: In this environment, the dealer deals several decks of cards. The player's goal is to correctly identify the suit of the card that was dealt $k$ steps ago. Observations at each timestep $t$ include the currently dealt card and an indicator of whether the action at $t-1$ successfully identified the suit from $k$ steps back. The player's action is to predict the suit of the card from $k$ steps ago. A positive reward is given for a correct prediction and a penalty for an incorrect one. The hyperparameter $k$ as well as the number of decks decides the memory length needed for this model. As $k$ increases, the complexity of the task correspondingly escalates.
    \item \emph{AutoEncode}: In this environment, the player receives a sequence of cards, and the goal is to output the cards in reverse order. The game consists of two phases: WATCH and PLAY. During the WATCH phase, a deck of cards is shown to the player, which they are expected to memorize. The observation during this phase is the currently shown card, with no action required from the player and zero reward. After the WATCH phase, the game switches to PLAY. In this phase, the player is prompted to recall and output the suit of the next card in the reverse order of presentation, based on their memory of the WATCH phase. The player's action is to predict the suit of the card in reverse order. Correct predictions are rewarded with $1/N$, and incorrect predictions incur a penalty of $-1/N$, where $N$ is the total number of cards shown in the WATCH phase. The complexity of the task depends on the number of decks.
\end{enumerate}

\paragraph{Games.}
\begin{enumerate}
    \item \emph{Minesweeper}: In this environment, players engage in the classic Minesweeper game. Players lack visibility of the board. Each observation per turn includes the count of mines in the adjacent cells, along with indicators to show whether the current grid cell has been visited or contains a mine. Players choose a grid cell to reveal as their action. The reward system is structured as follows: players receive a success reward for revealing an unexplored grid cell, and a penalty is incurred for revealing a cell with a mine or a cell that has already been visited.
    \item \emph{Concentration}: In this card-based environment, decks are shuffled and placed face-down. The player flips two cards at a time face up, receiving a reward if the flipped cards match. Once cards are flipped and matched, they remain face up. A positive reward is awarded if the two flipped cards are matched, either in rank or color, and have not been flipped before. A penalty is applied if the cards do not match or have been flipped previously.
    \item \emph{Battleship}: In the Battleship environment, players select locations to 'hit' on a grid. The game provides feedback with a reward or penalty based on whether a ship is hit, a shot is missed, or if the selected location has been previously targeted. A perfect strategy, where the player never misses, yields a total reward of 1. Conversely, a strategy of random guessing across all locations results in a total reward of 0.
\end{enumerate}

\paragraph{Sparse Reward.}
\begin{enumerate}
    \item \emph{Delayed Catch}: Delayed Catch is a variant of Catch, where an agent controls a paddle at the bottom of a $7\times 7$ grid, moving left or right to catch a falling ball. The ball drops from a random position at the top in a straight line. In Delayed Catch, all rewards are given at the end of the episode.
    \item \emph{Dark Key-to-Door}: The environment consists of a dark $9 \times 9$ room where the agent starts at a random location. It must first find an invisible key to receive a reward of 1, then locate an invisible door for another reward of 1. The maximum episode length is 50 steps.
\end{enumerate}

\subsection{Environment Modifications}
\paragraph{Observation Encoding in POPGym Environments}
In some POPGym environments, we have incorporated reward signals into observation data as discrete variables to measure prediction loss independent of error metrics ({\it i.e.,} all prediction errors evaluated in cross-entropy loss). Consequently, across all environments, we transform the continuous reward values into discrete categories. Specifically, for the Repeat Previous and Autoencode environments, we encode the reward as follows: `0' when the reward is 0, `1' indicates a positive reward, and `2' indicates a negative reward. For the Battleship environment, we do not include the reward signal separately since the hit/miss signal already conveys the information. For the Minesweeper environment, the reward is encoded as follows: `0' represents a positive reward, `1' indicates hitting a mine (thereby a negative reward), and `2' denotes a repeated tile.

\paragraph{Other Environments Implementations}
For sparse-reward environments, we use existing implementations: Dark Key-to-Door from the Amago repository\footnote{\url{https://github.com/UT-Austin-RPL/amago/blob/main/amago/envs/builtin/toy_gym.py}} and Delayed Catch from the POMDP-Baselines repository\footnote{\url{https://github.com/twni2016/pomdp-baselines/blob/main/envs/credit_assign/catch.py}}.

\subsection{Design Test Action}
In all our experiments, we pick one random timestep within each episode and take a designed test action when generating trajectories. In our experiments in RepeatPrevious, AutoEncode, and Mujoco, we sample a random action uniformly from the set of possible actions. In the Minesweeper and Battleship environment, we randomly choose one that has not been played yet. In Concentration, we choose one that has been played before. We mention that such a test design did not have a significant impact on the overall performance, though it slightly speeds up the convergence.

\section{Experimental Setup}\label{appendix:exp_results}
All experiments were conducted on an NVIDIA GeForce RTX 3090. 
In each episode, we collect the actions up to the maximum episode length of this environment. If the environment ends early, we will pad with dummy observation, actions and rewards.

\subsection{Model Specification}
\label{appendix:model}
Our approach employs an actor-critic framework, with a Q-network for value estimation and a policy network for action selection. The Q-network follows a double Q-learning structure, where each Q-network is a 2-layer fully connected network (FCN) with Tanh activations. The policy network shares the same 2-layer FCN architecture with Tanh activations.

For state, action, and reward projections into hidden states, we use an \texttt{Embedding} layer for discrete inputs and a \texttt{Linear} layer for continuous ones.
For future prediction, we adopt a simple GRU model with one layer and a hidden dimension of 16 to capture temporal dependencies.

\subsection{Training Hyperparameters}
The following hyperparameters are shared across environments.
\begin{itemize}
    \item For the SAC agent learning
    \begin{enumerate}
        \item learning rate for the actor model: 0.0001 
        \item learning rate for the critic model: 0.0002
        \item discount factor $\gamma$: 0.99
        \item entropy regularizer: 0.01
    \end{enumerate}
    \item For the future prediction models
    \begin{enumerate}
        \item embedding dimension: 128 for MuJoCo and GridWorld, 256 for POPGym
        \item number of layers: 3
        \item number of heads: 4
    \end{enumerate}
    \item For the PSR training
    \begin{enumerate}
        \item learning rate for sequential model: 5e-5
        \item weight decay: 0.0001
        \item buffer size: 30,000
    \end{enumerate}
\end{itemize}
One special case is the Concentration environment in POPGym, where the number of states is either 104 or 52, corresponding to the number of decks. As mentioned earlier, we evenly split the embedding table across state dimensions. To better align with the 256-dimensional embedding, we set the embedding dimension to 260 for ConcentrationEasy/Hard and 208 for ConcentrationMedium.

Other environment-specific hyperparameters are summarized in Table~\ref{table:hyper_param_table}. In each step, we first collect $T_{gen}$ steps, then update PSR for $T_{psr}$ steps, followed by an RL update for $T_{rl}$ steps.  The ratio between $T_{psr}$ and $T_{rl}$ reflects the step sizes $\alpha$ and $\beta$ as defined in Algorithm~\ref{algo:interleaving_PSR}.

\subsection{Amago Hyperparameters}
For the Delayed Catch and Dark Key-to-Door environments, we use the Amago implementation with its default settings, tuning only the $\alpha:\beta$ ratio by adjusting the weight of the future prediction loss in the overall loss. We also list out the hyperparameter they use in Table~\ref{table:hyper_param_table}: there is no burn-in steps, and the trajectory generation is in timesteps scale instead of episodes.

\begin{table}[h]
\centering
\resizebox{0.95\textwidth}{!}{
\begin{tabular}{lccccc}
    \toprule
    \multirow{2}{*}{} & \multirow{2}{*}{\shortstack{History Summarizing \\ Model Backbone}} & \multirow{2}{*}{Burn-in Steps} &  \multirow{2}{*}{\shortstack{Traj Generation \\ Steps $T_{gen}$}} & \multirow{2}{*}{PSR steps : RL steps} \\
    \\
    \midrule
    GridWorld & GPT & 5000 & 10 & 50 : 500 \\
    AutoEncode & Amago-GPT & 10000 & 2 & 200 : 500 \\
    Battleship & GPT & 10000 & 5 & 50 : 500 \\
    Concentration & Amago-GPT & 10000 & 10 & 50 : 500 \\
    Minesweeper & GPT & 10000 & 10 & 50 : 500 \\
    RepeatPrevious & Amago-GPT & 4000 & 2 & 200 : 500 \\
    MuJoCo & GPT & 5000 & 2 & 1 : 5 \\
    \bottomrule
    \toprule
    \multirow{2}{*}{Amago Implementation} & \multirow{2}{*}{\shortstack{History Summarizing \\ Model Backbone}} & \multirow{2}{*}{Burn-in Steps} &  \multirow{2}{*}{\shortstack{Traj Generation \\ Timesteps}} & \multirow{2}{*}{$\alpha : \beta$} \\
    \\
    \midrule
    Delayed Catch & Amago-GPT & 0 & 1000 & 0.03 : 1 \\
    Dark Key-to-Door & Amago-GPT & 0 & 1000 & 10 : 1 \\
    \bottomrule
\end{tabular}
}
\vspace{-0.8em}
\caption{\emph{Hyper-Parameter Table}}
\label{table:hyper_param_table}
\end{table}

\section{Additional Results}
\subsection{Shorter-Term Memory Benchmarks} \label{subsec:shot_term_memory}
We conduct additional analysis on benchmarks where short-term memory is sufficient for capturing historical contexts—verified when the simplest \eename~trained with a GRU summarizer already performs well. We study continuous control tasks from MuJoCo~\citep{todorov2012mujoco}, incorporating the partial observability modifications from~\citep{meng2021memory, ni2022recurrent}. Following \citep{ni2022recurrent}, we introduce partial observability by omitting certain observations and adding noise to the remaining ones\footnote{\url{https://github.com/twni2016/pomdp-baselines/blob/main/envs/pomdp/wrappers.py}}. We evaluate the performance of \eename~with GRU, \eename~with TF, and \algname~with TF on four control tasks: Ant, Cheetah, Hopper, and Walker. 

\begin{figure}[ht!]
    \centering
    \ifarxiv
    \includegraphics[scale=0.6]{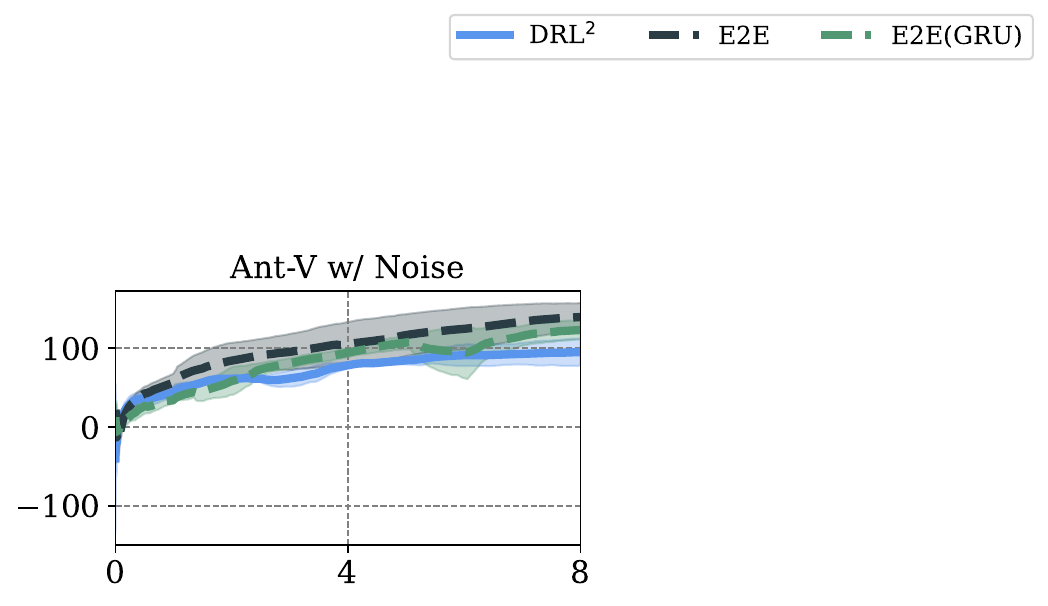}
    \else
    \includegraphics[scale=0.4]{Figures/mujoco_legend.pdf}
    \fi
    \hspace{500mm}
    \includegraphics[height=0.14\textwidth]{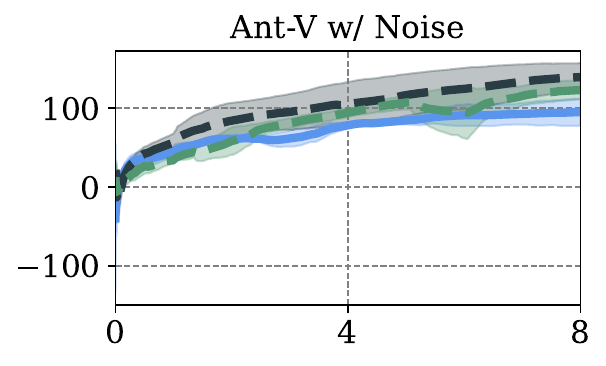}
    \includegraphics[height=0.14\textwidth]{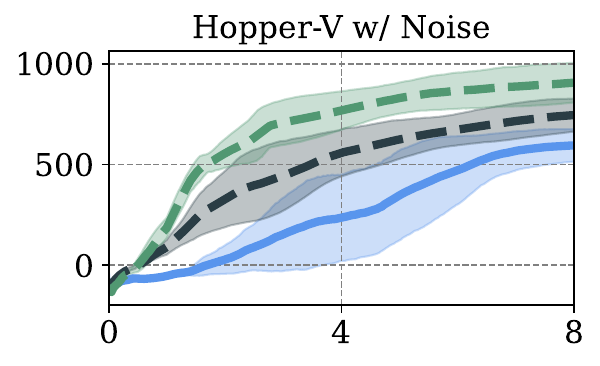}
    \includegraphics[height=0.14\textwidth]{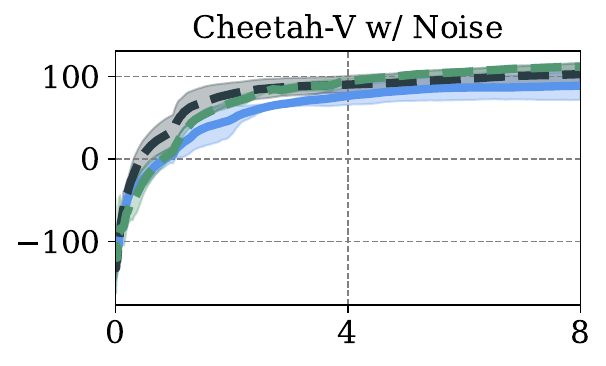}
    \includegraphics[height=0.14\textwidth]{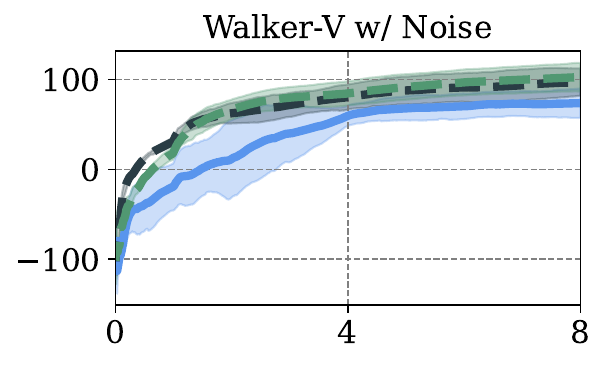}
    \caption{\small \emph{Performance Comparison of \algname\ and \eename\ Methods in Noisy MuJoCo Environments.} In these MuJoCo environments where E2E-GRU performs well, PSR-TF slightly underperforms relative to E2E-TF.}
    \label{fig:mujoco}
\end{figure}

The results are presented in Fig.~\ref{fig:mujoco}. In summary, when short-term memory suffices to resolve partial observability, we observe that the \eename~training, especially with the simplest GRU architecture, is superior to the future prediction approach that incorporates auxiliary prediction losses, which may not be best aligned with the original objective. This aligns well with observations from \citet{ni2022recurrent, grigsby2023amago} where end-to-end training is often a strong baseline in many partially observable environments with shorter-term memories. Nevertheless, the induced representation independently by future prediction can still be used to achieve high rewards, albeit slower convergence to the optimal policy, as we can observe in Fig.~\ref{fig:mujoco}.

\subsection{Full Result in POPGym}
\begin{figure*}[ht!]
    \centering
    \begin{tabular}{ccc}
        \includegraphics[height=0.16\linewidth]{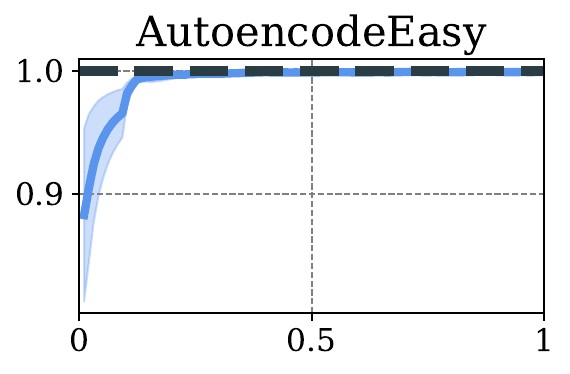} & 
        \includegraphics[height=0.16\linewidth]{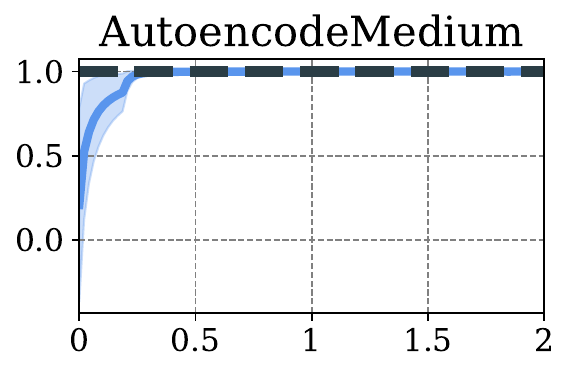} & 
        \includegraphics[height=0.16\linewidth]{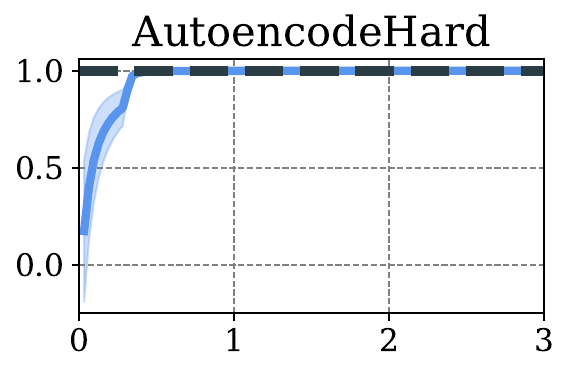} \\
        \includegraphics[height=0.16\linewidth]{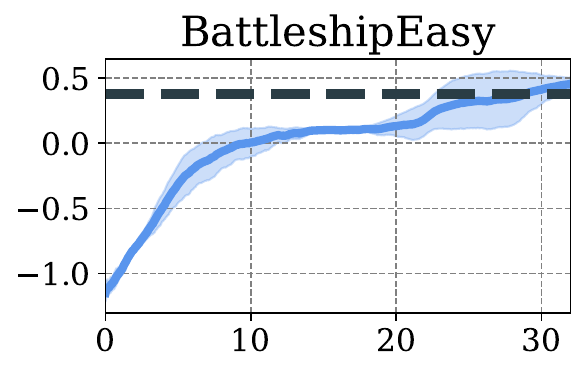} & 
        \includegraphics[height=0.16\linewidth]{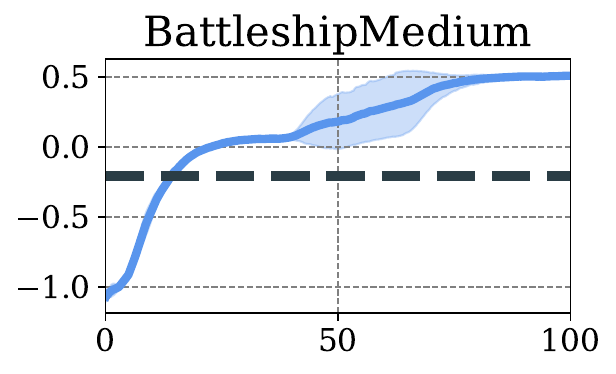} &
        \includegraphics[height=0.16\linewidth]{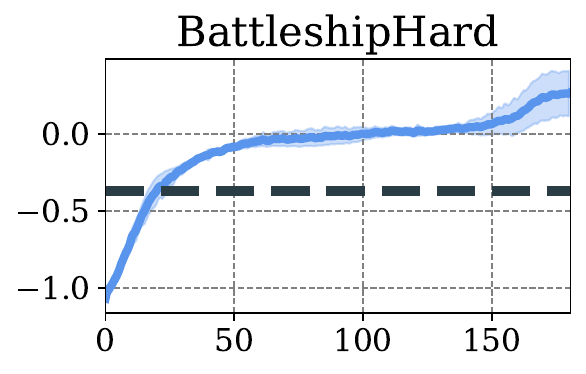} \\
        \includegraphics[height=0.16\linewidth]{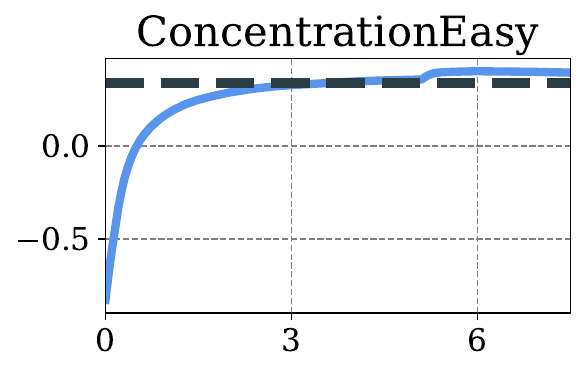} &
        \includegraphics[height=0.16\linewidth]{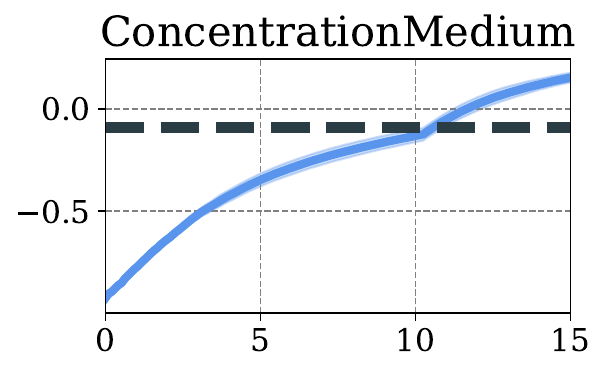} &
        \includegraphics[height=0.16\linewidth]{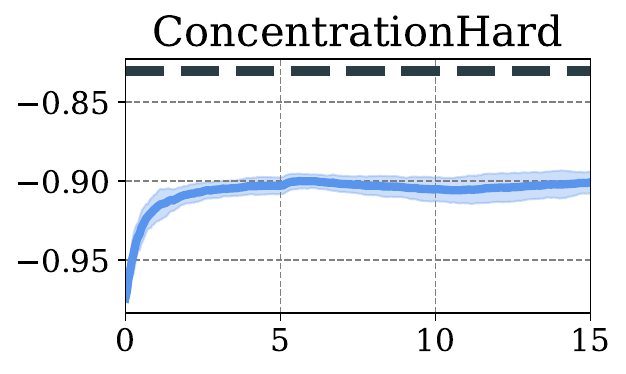} \\
        \includegraphics[height=0.16\linewidth]{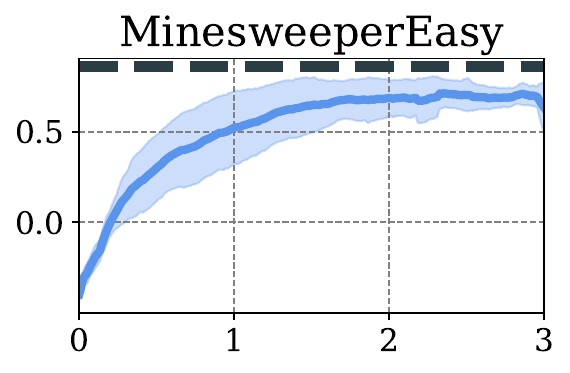} &
        \includegraphics[height=0.16\linewidth]{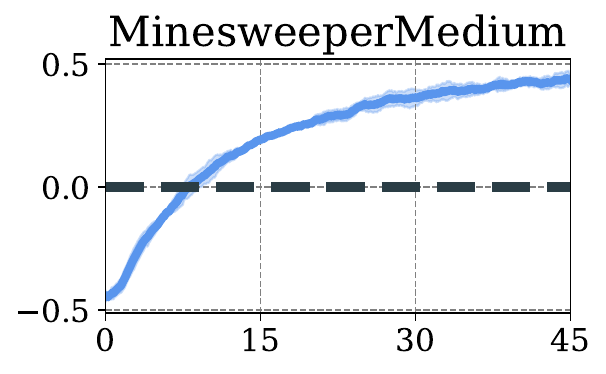} & 
        \includegraphics[height=0.16\linewidth]{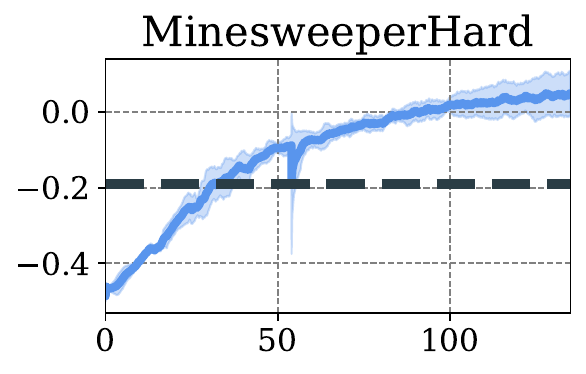} \\
        \includegraphics[height=0.16\linewidth]{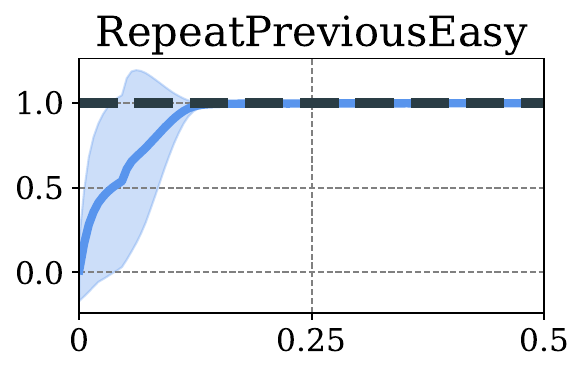} &
        \includegraphics[height=0.16\linewidth]{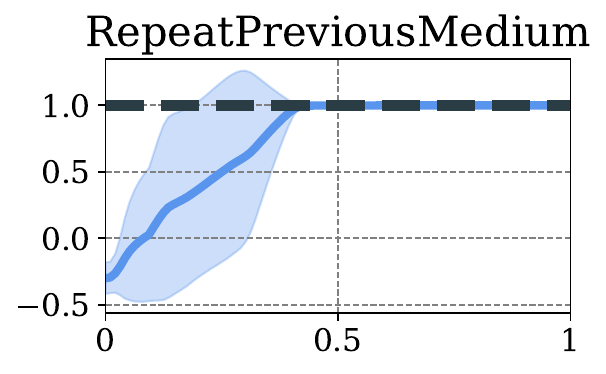} &
        \includegraphics[height=0.16\linewidth]{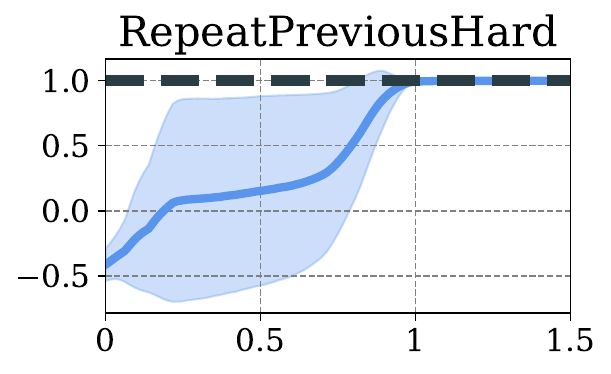} \\
    \end{tabular}
    \caption{\small Performance of \algname on POPGym benchmark. The dashed line represents the best reported on-policy or off-policy method. The x-axis shows timesteps on a million scale.
    }
    \label{fig:all_results}
\end{figure*}

Figure~\ref{fig:all_results} displays the learning curves from multiple runs, with the dashed lines representing the performance of the reported state-of-the-art (SOTA) on-policy or off-policy methods. 
In each environment, our algorithm, \algname, successfully learned policies that performed above the chance level. The configuration of backbone sequence models is decided by the complexity of future prediction tasks in each environment. %

